\documentclass[a4paper,conference]{IEEEtran}
\usepackage{graphicx}
\usepackage{hyperref}
\usepackage{float}
\usepackage{makecell}
\usepackage{longtable}
\usepackage{listings}
\usepackage{amssymb}
\usepackage[font=small,labelfont=bf]{caption}
\usepackage{subcaption}
\usepackage{booktabs}
\usepackage{diagbox}
\usepackage{amsmath}
\usepackage{tabularx,ragged2e}
\ifCLASSINFOpdf

\else

\fi

\begin{document}

\title{LaHAR: \underline{La}tent \underline{H}uman \underline{A}ctivity \underline{R}ecognition using LDA}

\author{
\IEEEauthorblockN{Zeyd Boukhers}
\IEEEauthorblockA{University of Koblenz-Landau\\
boukhers@uni-koblenz.de}
\and

\IEEEauthorblockN{Danniene Wete}
\IEEEauthorblockA{University of Koblenz-Landau\\
dwete@uni-koblenz.de}

\and
\IEEEauthorblockN{Steffen Staab}
\IEEEauthorblockA{University of Stuttgart\\
steffen.staab@ipvs.uni-stuttgart.de}}

\maketitle

\begin{abstract}
Processing sequential multi-sensor data becomes important in many tasks due to the dramatic increase in the availability of sensors that can acquire sequential data over time. Human Activity Recognition (HAR) is one of the fields which are actively benefiting from this availability. Unlike most of the approaches addressing HAR by considering predefined activity classes, this paper proposes a novel approach to discover the latent HAR patterns in sequential data. To this end, we employed Latent Dirichlet Allocation (LDA), which is initially a topic modelling approach used in text analysis. To make the data suitable for LDA, we extract the so-called ``sensory words'' from the sequential data. We carried out experiments on a challenging HAR dataset, demonstrating that LDA is capable of uncovering underlying structures in sequential data, which provide a human-understandable representation of the data. The extrinsic evaluations reveal that LDA is capable of accurately clustering HAR data sequences compared to the labelled activities. 
\end{abstract}


\IEEEpeerreviewmaketitle

\section{Introduction}
The recent expansion of smartphones and portable devices endowed with sensors and the emergence of new technological advances such as the Internet of Things (IoT) has led to the pervasiveness of the use of sensors. It is estimated that by the early 2020s, their number will have already surpassed one trillion \cite{ebner2015}. This massive amount of available sensors generate a massive and continuously increasing amount of sequential data every day.

Recently, several studies have been conducted to classify human activities using sequential multi-sensor data \cite{shirahama2017, mcnames2019two, badawi2018multimodal, iosa2016wearable}. Most of these approaches are supervised and focus on categorizing data into predefined activities~\cite{bao2004, anguita2013public, shirahama2017, badawi2018multimodal}. In real scenarios, the defined activity classes are suitable only for a specific Human Activity Recognition (HAR) task and a new annotation process has to be conducted for different HAR tasks. For example, activity classes needed for an elderly monitoring system are different from those needed for fitness tracking.


Furthermore, most of the classical approaches assume that only one activity is performed at a time~\cite{anguita2013public, badawi2018multimodal, iosa2016wearable}. In reality, an activity can embed other activities. For example, teeth brushing needs other basic activities to be performed such as standing/sitting, folding the arm and moving the forearm back and forth. Other approaches consider different levels of activities (mostly two), where activities of a higher level can embed those of lower-levels~\cite{huynh2008discovery}. The data sequences are first classified based on low-level activities, which are used afterwards to classify the data sequence into higher-level activities. This means that activities at the same level are completely independent and cannot be performed simultaneously. Finding the best hierarchy that is adequate for every conceivable activity is difficult to achieve. Also, this hierarchy makes the designed model limited in terms of the scalability to consider new activities and the generalizability on other tasks. These problems can be overcome by considering that more than one activity is performed in a bounded time span and then depending on the HAR task, the main activity is derived. This means that a data sequence is considered to comprise a mixture of activities.

Probabilistic topic models are powerful unsupervised methods used in Natural Language Processing (NLP). These models are applied mainly to cluster large corpora of documents by finding the main topics that prevail in the documents. Latent Dirichlet Allocation (LDA) is one of the most popular probabilistic topic models. LDA assumes that each document consists of a mixture of a fixed number of topics, and topics are distributions over words \cite{blei2003}. This assumption coincides with our consideration of a data sequence as a mixture of activities. Therefore, this paper seeks to tackle the HAR problem by treating it as a topic modelling task and applying LDA for its analysis. We assume that LDA can extract the mixture of activities from the sequences in the same way it extracts topic mixtures from text documents. LDA is a bag-of-words (BoW)-based model in which documents are treated as a set of discrete words without considering their order. This is a big barrier to applying LDA on multisensor data, which are sequences of continuous numerical values. To handle this, this paper extract from these sequences a set of discrete words called ``\emph{sensory words}'', which LDA can use to predict latent activities. The main contributions of this paper are: 
\begin{itemize}

\item We introduce a new representation of multisensor data by extracting discrete patterns called ``sensory word''.
\item We treat a data sequence as a mixture of activities, which can give further information about the sequence instead of only its activity class. 
\item We model the distribution of activities using LDA.
\item The experimental results on a HAR dataset demonstrate the effectiveness of our LaHAR to cluster data sequences based on the performed activities.
\end{itemize}

Following this section, Section~\ref{related} discusses the related works. Section~\ref{approach} presents the proposed approach and Section~\ref{exp} presents the conducted experiments and the obtained results that validate the effectiveness of our approach. Finally, Section~\ref{conclusion} concludes this paper and gives insight into future directions.

\section{Related Work}
\label{related}
In this section, we review previous works that were concerned with applying topic models on non-textual data. We also discuss related contributions to human activity recognition using multi-sensor data.\\

\subsection{Topic models for non-textual data}
Topic models have been widely explored for clustering, especially on text documents \cite{blei2003, hofmann2001unsupervised}. However, researchers have also examined the application of topic models to non-textual data such as images~\cite{espinoza2017, rasiwasia2013latent}, videos~\cite{gong2019dynamic, varadarajan2017active}, time series~\cite{emonet2013temporal} and audio signals~\cite{kim2012latent, kim2009acoustic} etc.
 In~\cite{zou2014}, Zou et al. applied LDA to cluster trajectories obtained from videos of crowded scenes. They extracted visual words from the video scenes and then fed into LDA. In \cite{bahmanyar2018multisensor}, a multimodal Latent Dirichlet Allocation Model (mmLDA) was developed to perform land-cover classification based on Multisensor Earth Observation Image (MEOI). In this method, image patches are represented by bag-of-words, which in return are fed to the mmLDA. Furthermore, topic models have also found successful application on time series sensory datasets. for example, \cite{puschmann2017using} assumed that there are structures and relations in heterogeneous smart city data streams which are not visible in the raw data. The authors used LDA to uncover the hidden structures and relations in the smart-city data stream. They employed the symbolic aggregate approximation (SAX) method~\cite{lin2003symbolic} in order to convert raw sensor data into string-based patterns. These string-based patterns are aggregated into ``virtual documents'', which are then represented as BoW to be fed into LDA. \cite{kim2009acoustic, imoto2018acoustic}, on the other hand, extracted acoustic words from audio signal recorded by audio sensors. This representation of the audio signal allowed LDA to learn hidden acoustic topics in a given audio signal in an unsupervised way. \cite{raanan2018detection} developed an online version of LDA to cluster state-sensor data from autonomous underwater vehicles (AUVs) in order to perform automatic fault diagnosis for AUVs based on the topic model outputs.

\subsection{Human Activity Recognition using multi-sensor data}
There is a lot of research addressing HAR using multi-sensor data \cite{bao2004, badawi2018multimodal, anguita2013public, shirahama2017}. However, HAR is mostly envisaged as a supervised machine learning problem. Although classification algorithms such as SVM yield promising results for HAR \cite{anguita2013public, shirahama2017}, these algorithms often rely on handcrafted features that are designed based on prior knowledge and manual investigation. In addition, the extracted handcrafted features are often incapable of handling compound activities especially with the present flood of multimodal and high dimensional sensory data \cite{nweke2018deep}. One solution to overcome the shortcoming of handcrafted features is to use the codebook method, which is a feature learning approach that extracts useful features from a large amount of data by applying clustering on the raw data \cite{shirahama2017}. More recently, deep learning methods have evolved as powerful techniques for feature extraction and classification of simple and complex human activity recognition in mobile and wearable sensors. For example, the convolutional neural network (ConvNet) was proposed for activity recognition in \cite{ronao2016human} and \cite{jiang2015human}. Another popular deep learning approach of long short-term memory (LSTM) was leveraged to perform HAR using mobile devices in \cite{singh2017human, zhao2018deep, yu2018multi}. However, the deep learning-based approaches are often affected by noise and the limited size of data, leading to an unsatisfactory performance \cite{chen2018distilling}.

Besides, owing to the growing amount of sensory data and the difficulty to obtain annotated training examples, clustering approaches have increasingly become useful to investigate HAR in an unsupervised manner \cite{jothi2018clustering}, for instance, explored three clustering algorithms: K-means, hierarchical agglomerative clustering and Fuzzy C-means (FCM) for human activity recognition. Experimental results carried on a time series dataset from smart devices have shown that FCM algorithm effectively categorizes the activities. However, FCM clustering, like other conventional clustering algorithms has problems with high dimensional datasets \cite{winkler2012problems} and can therefore only handle time series with a short length. \cite{he2018wavelet} seeks to address this problem by proposing a wavelet tensor fuzzy clustering scheme (WTFCS) for multi-sensor activity
recognition. Feature tensors of multiple activity signals are created using the
discrete wavelet packet transform (DWPT). Then, a Multilinear Principal Component Analysis (MPCA) is leveraged to reduce the dimensionality of feature tensors. Based on the principal feature initialization and the tensor fuzzy membership, a new fuzzy clustering is developed to identify different activity feature tensor groups. The obtained results reveal that this approach works better than FCM and fuzzy clustering with tensor distance. However, the authors reported that this approach does not perform well with complex activities.
\subsection{Topic models for HAR analysis}
Researchers have also tackled HAR using probabilistic topic models. For example, Merino and Atzmueller~\cite{merino2018behavioral} studied driving behaviour by applying topic models on in-vehicle sensor data. The authors utilized the Symbolic Aggregate Approximation (SAX) method to process driving data and convert them into SAX words. This representation was supplied to  LDA to infer behavioural driving patterns. Steil and Bulling, \cite{steil2015discovery} applied LDA to discover users’ everyday activities from their long-term visual behaviour. Their method encodes visual behaviour like saccades, fixations, and blinks into a string sequence from which a bag-of-words representation is generated to fit the LDA model.\\

The recognition of complex activities using topic models was studied in \cite{huynh2008discovery}. The authors consider daily routines (commuting, office work etc.) as higher-level activities that embed lower-level activities such as walking, using the phone, discussing at a whiteboard, etc. LDA was used to model daily routines as probabilistic combinations of multiple activities patterns. In their approach, the authors applied the LDA on sensory data using two different methods. In the first method, supervised learning is used to assign activity labels to the sensor data instance. These labels are then utilized as words to identify complex activity patterns in an unsupervised fashion with LDA. The second method was completely unsupervised. In this method, clustering is used to generate a vocabulary of labels, which are then fed as words to the LDA model. The evaluation was conducted using data recorded about the daily life of one subject over a period of 16 weekdays using two wearable sensors. \\

In contrast to \cite{huynh2008discovery}, we consider that short time activities can be also viewed as a combination of multiple activity patterns. Further, we do not build a hierarchy level between activities. Instead, the raw subsequences values are used as features for clustering.

\section{Latent Human Activity Recognition (LaHAR)}
\label{approach}
\begin{figure*}
 \center
        {\includegraphics[width=\textwidth]
       {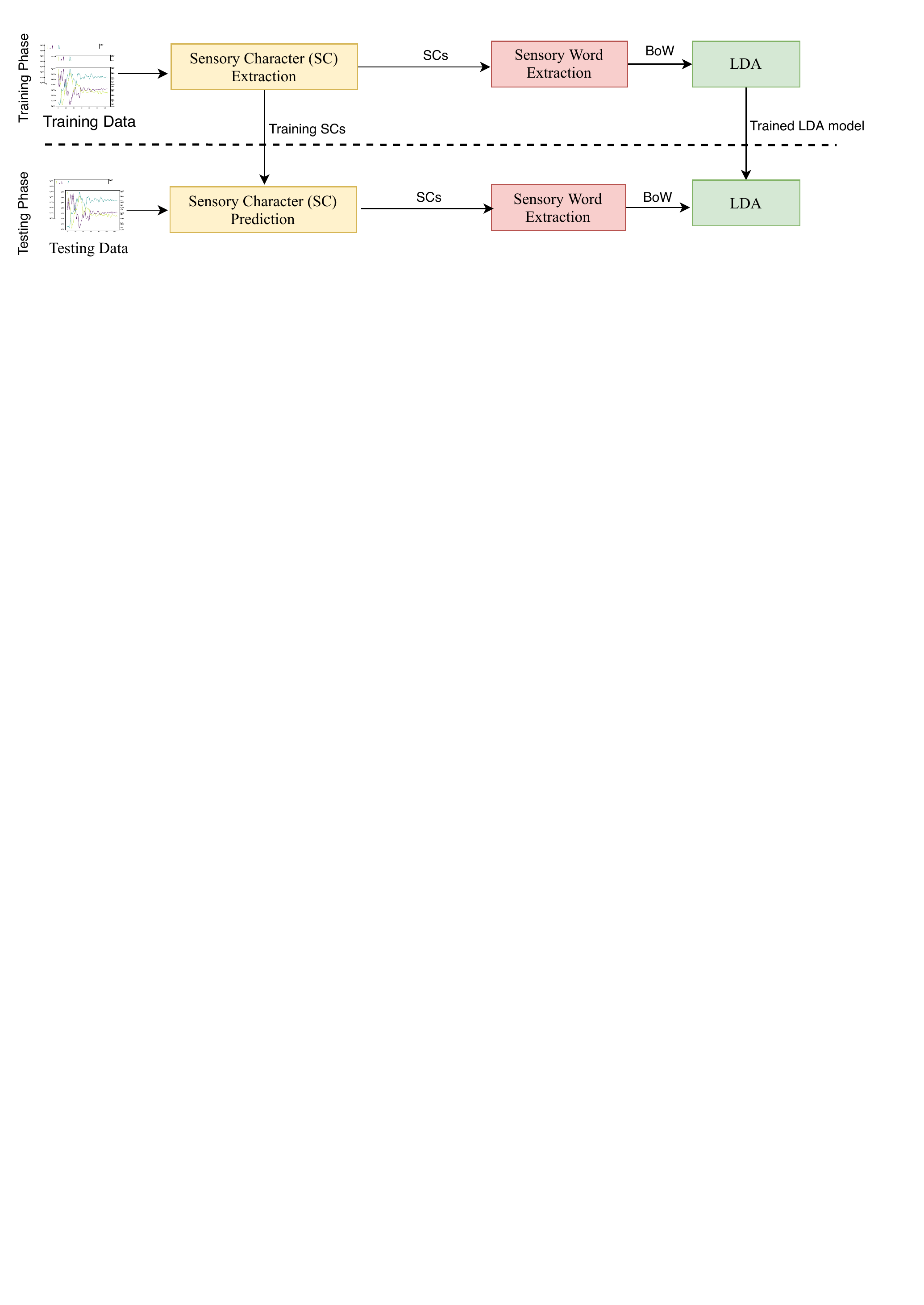}}
       \vspace{-8in}
       \caption{\label{fig:ldaPipe} Procedure of extracting \textit{sensory words} from sensory data.}
\end{figure*}

In HAR, most of the employed sensors acquire data on multiple channels. For instance, an accelerometer sensor on a smartphone provides measurements on three different channels: $(x, y, z)$. The measurements are generated along all channels simultaneously. 
Formally, let us defined $\mathbf{S}=\{\mathbf{s}_{m,l}\}_{l=1}^{L}, m=1,\cdots,M$ be multisensor data sequence where $\mathbf{s}_{m,l}$ are the sequential measurements of the $m$th channel in the $l$th sensor. Supposing that each $\mathbf{s}_{m,l}=[s_1,s_2,\cdots,s_t]$ is of length $t$ that is equal for all $M$ channels in the $l$th sensor. Let  $D=\{\mathbf{S}_1,\mathbf{S}_2,\cdots,\mathbf{S}_N\}$ be a multisensor dataset composed of $N$ data sequences having the same number of sensors and channels. In this paper, a data sequence denotes a segment of measurements of all channels in all sensors in a bounded time span that is supposed to embed one activity in the classical concept of HAR. The flowchart of our approach to model activity-sensory word distribution is shown in Figure \ref{fig:ldaPipe}. The process consists of three steps, which are represented by rectangles. The first step is sensory character extraction which collects numerous subsequences from raw sensor sequences. Afterwards, for each channel, it clusters the extracted subsequences into a finite number of groups called sensory characters. For unseen data sequences, the sensory characters obtained in the training phase are assigned to the extracted subsequences based on the squared Euclidean distance. The second step extracts the sensory words of each data sequence to generate BoW representation of the instance. Lastly, LDA is applied on the BoWs obtained from data sequences. The trained LDA model is applied to unseen data sequences without update or retraining. We will describe each step in details in the subsequent subsections. The output of each step is indicated with an arrow. 
 
\subsection{Sensory characters extraction}
\begin{figure*}
\center
        {\includegraphics[width=\textwidth]
       {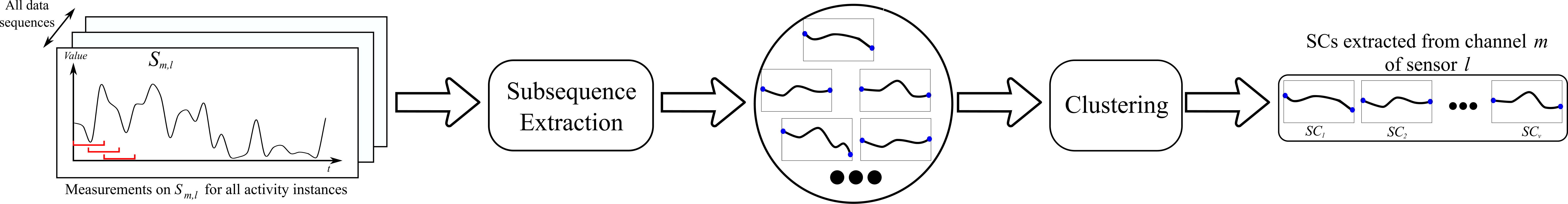}}
       \caption{\label{fig:sen_char} An illustration of extracting sensory characters from all sequential measurements corresponding to a single channel.}
       \vspace{-0.3cm}
\end{figure*}

In this step, we separate sensors and process each channel independently. Figure~\ref{fig:sen_char} illustrates the extraction of sensory character which starts by successively extracting subsequences from each channel using the sliding window technique; Given a sequence $\mathbf{s}$ of length $t$ and a sliding window size $p$, $l'$ subsequences $\{\mathbf{q}_1,\mathbf{q}_2,\cdots,\mathbf{q}_l'\}$ are extracted by shifting the sliding window across $\mathbf{s}$. Here, a subsequence $\mathbf{q}$ of $\mathbf{s}$ is a sampling of length $p \leq t$. The shifting of the sliding window is applied with 50\% overlap so that the half of data points from the previous subsequence are included into the current one and so on.

Since the abundant subsequences which are collected from each channel are likely to be similar to each other, clustering is applied to obtain a finite set of prominent subsequences that are statistically distinctive. Specifically, we used K-means to independently cluster the subsequences extracted from every channel. For a channel $m$ in sensor $l$, the k-means algorithm constructs $v$ clusters, which we call sensory characters (SCs). Note that since the measurements in all channels have the same length $t$, $v$ is equal for all channels.

\subsection{\textit{Sensory word} extraction (SWE)}\label{swe}
\begin{figure}
\center

        {\includegraphics[width=0.95\linewidth]
       {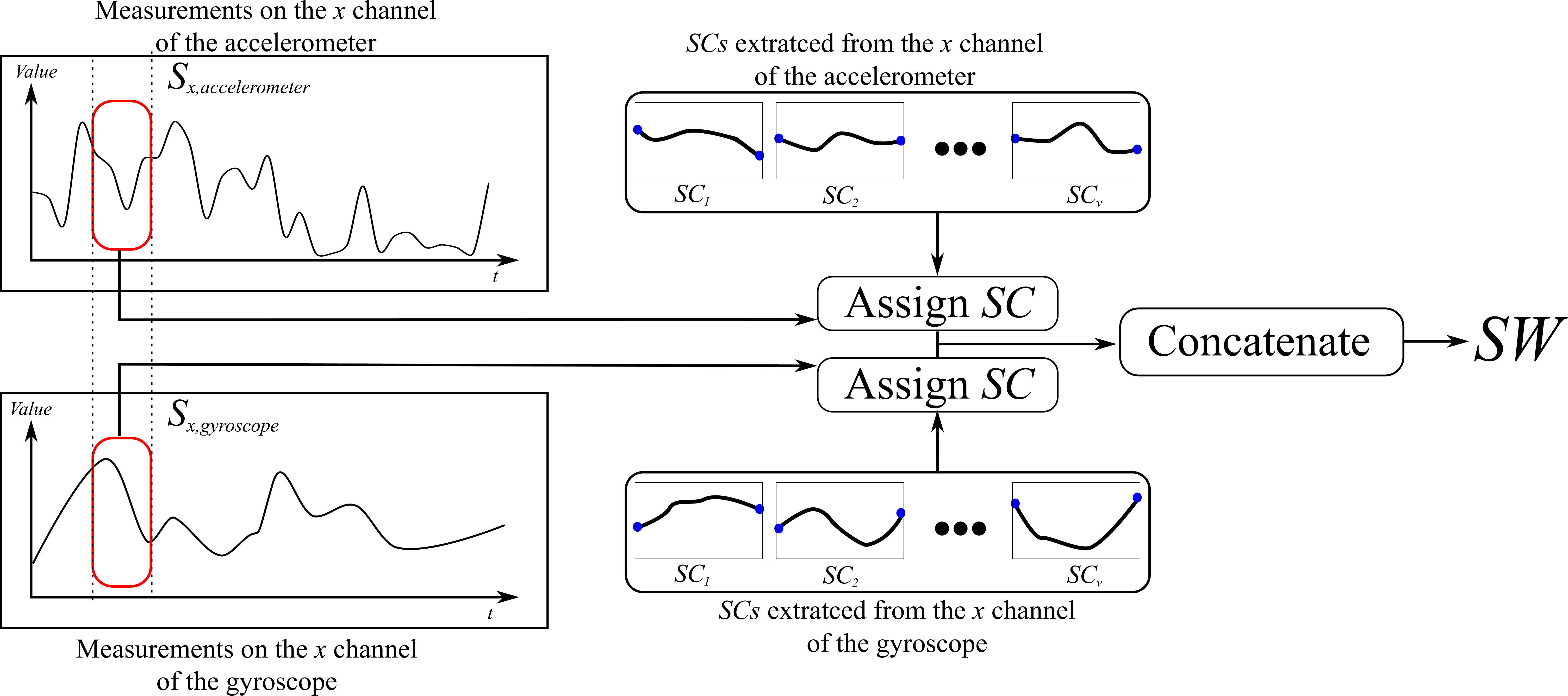}}
       \caption{\label{fig:sen_word} An illustration of the composition of sensory words from sensory characters.}
       \vspace{-0.3cm}
\end{figure}

In this step, we use the obtained sensory characters to build a bag of sensory words $W=\{w_1, w_2, \cdots, w_{|d|}\}$ for each data sequence $d$, where $|d|$ is the number of words in $d$. It is important to note that all the channels of every smartphone sensor used in HAR express a certain motion (e.g. acceleration, rotation) of the smartphone on three axes $(x, y, z)$ corresponding to abscissa, ordinate and applicate of the smartphone, respectively. Therefore, we assume that the motions captured by different sensors on the same axis are correlated and consequently a sensory word is composed by aggregating the sensory characters of all sensors on the same axis. Figure~\ref{fig:sen_word} illustrates the extraction of sensory words from a data sequence consisting of sequential measurements from two sensors; namely, accelerometer ($a$) and gyroscope ($g$). Let us consider the two synchronized subsequences $q_{x,a}$ and $q_{x,g}$ extracted from the $x$ channels of $a$ and $g$, respectively. First, each subsequence is assigned to its corresponding sensory character. More precisely, $q_{x,a}$ is assigned to the closest sensory characters among the $v$ characters obtained from the channel $x$ of $a$. Similarly, $q_{x,g}$ is assigned to its corresponding sensory character. Afterwards, the sensory word is formed by concatenating the sensory characters. This process is applied to all channels. Note that, the sensory words are associated with the channel label in order to distinguish between them. Consequently, among $V=(M \times v!)/\left(L! \times (v-L)!\right)$ words in the vocabulary, ($2\times t\times M)/p$ sensory words can be extracted from a single data sequence.

\subsection{Latent Dirichlet Allocation}

We use LDA with Gibbs sampling for inference to uncover activity classes in a set of data sequences that are represented by a bag of Sensory words. This is done by finding the latent variables that describe the relations between sensory words and the data sequence. The basic concept of LDA is illustrated in Figure~\ref{fig:lda_graph}. Given a collection of $D$ sensory instances, it is assumed that each data sequence $d$ is generated from a random mixture of latent components (activity classes) and each topic is a distribution over the $V$ words of the vocabulary. The generative process of LDA is as follows: 

\begin{figure}
\center
        {\includegraphics[width=6cm]
       {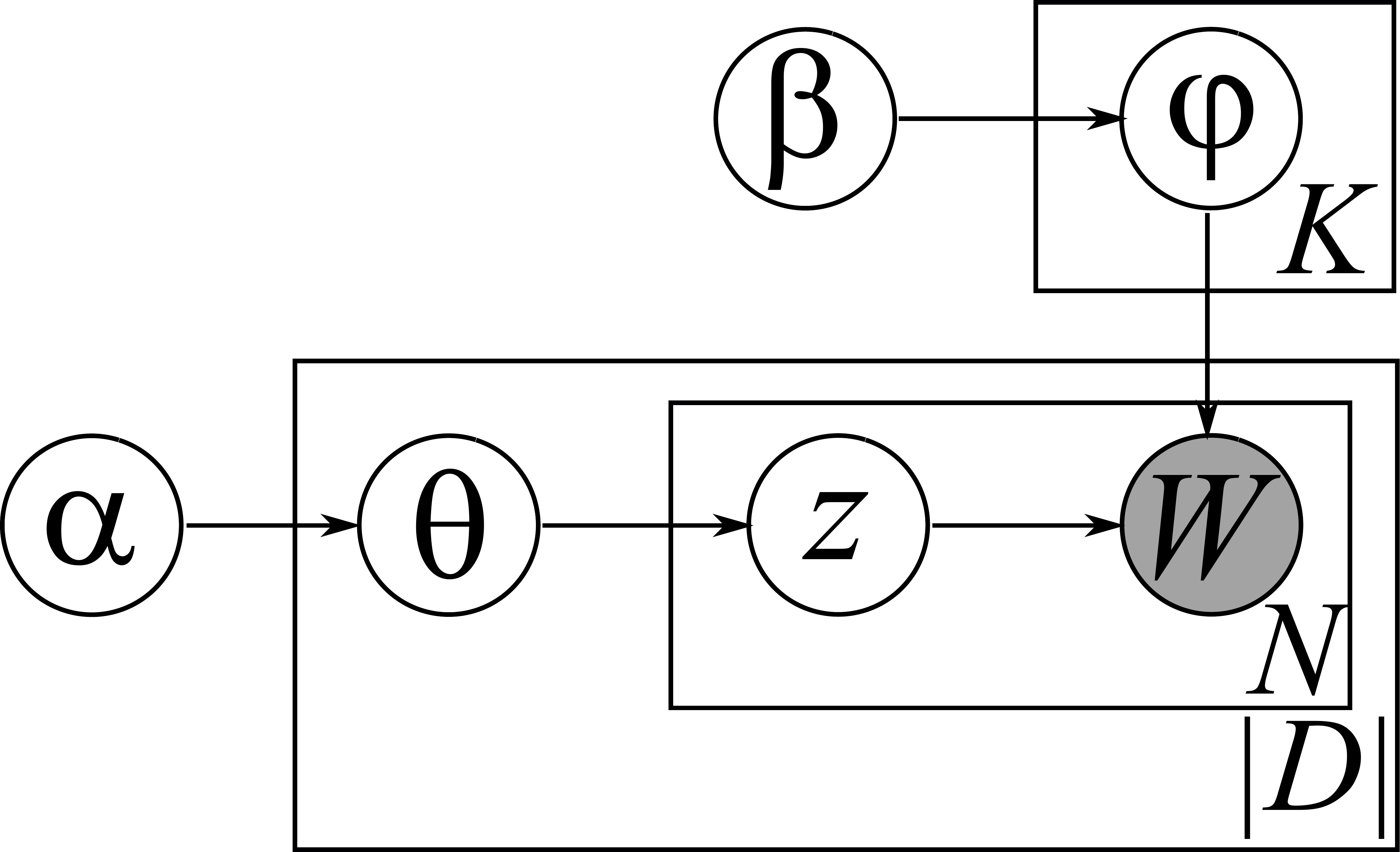}}
       \caption{\label{fig:lda_graph} Graphical representation of LDA topic model.}
       \vspace{-0.4cm}
\end{figure}

\begin{itemize}
    \item For each activity class $k \in K$:
    \begin{itemize}
      \item $\varphi_k  \thicksim Dirichlet(\beta)$
    \end{itemize}
  \item For each data sequence $d \in D$:
    \begin{itemize}
      \item $\theta_d  \thicksim Dirichlet(\alpha)$
      
      \item For each sensory word $w_i \in d$
          \begin{itemize}
              \item $z_i  \thicksim Multinomial(\theta)$
              \item $w_{i}   \thicksim Multinomial(\varphi_{z_i})$
        \end{itemize}
    \end{itemize}
\end{itemize}

where $\alpha$ and $\beta$ are the prior Dirichlet coefficients.
 
LDA assumes that all the above distributions are hidden and only the bag of sensory words of each data sequence is observed. The prior coefficients are estimated based on the maximum likelihood of the given data, where the number of activity classes is assumed to be known. In the evaluation, the number of activity classes is set to the number of classes provided in the ground truth. The task of LDA remains to learn the hidden distributions by reversing the generative process and expressing the conditional posterior distribution of the latent variables given the observed data:

\begin{equation}
    p(\theta,\varphi|\mathbf{w},\alpha,\beta)=\frac{p(\theta,\varphi,\mathbf{w}|\alpha,\beta)}{p(\mathbf{w}|\alpha,\beta)},
\end{equation}

where $\mathbf{w}$ is the bag of sensory words in the entire dataset.

\section{Experimental Results}
\label{exp}
The effectiveness of LaHAR is evaluated for its capability to discover the latent activities in a HAR dataset. This section provides a description of the used dataset and explains the implementation details. Afterwards, it presents and analyses the obtained results.
\subsection{Data set description and implementation details}

The conducted experiments are carried out on the UCI HAR data set \cite{anguita2013public}. In this dataset, a group of 30 volunteers performed Activities of Daily Living (ADL) while carrying a waist-mounted smartphone with integrated inertial sensors, namely accelerometer and gyroscope. These sensors provide measurements of length 128 data points at a constant frequency of 50 Hz on three channels. Each of the sequential activities is associated with an activity label (basic: SITTING = SI, STANDING = ST and LAYING = LA or dynamic: WALKING = WA, WALKING\_UPSTAIRS = WU and WALKING\_DOWNSTAIRS = WD. The data set is preprocessed by a noise filter to eliminate the components of gravitational acceleration. Note that we do not apply further preprocessing on the data. 

Our method is structured into training and testing phase using 7352 data sequences for training and 2947 data sequences for testing. The LDA model is trained in an unsupervised way without using the activity labels and is stored together with the sensory words and sensory characters to model the activity classes in the testing set.

 \subsection{Evaluation method}\label{evame}
 For evaluating our results, we use the ground truth activity labels, where the number of activity classes to be found by LDA is similar to the number of labels. Since for each data sequence, LDA learns the posterior probability distributions over activity classes, the data sequence is assigned to the activity class with the highest posterior. The class label $z$ of a data sequence $d$ was selected as:

\begin{center}
  $z$ = $\underset{k}{argmax(\theta_{d})}$  
\end{center}
where $\theta_{d}$ is the $K$-dimensional vector of activity distributions
for an instance $d$. According to the number of instances assigned to the activity classes obtained by LDA, the classes are mapped to the ground truth activity classes. We first map the topic to the class with the highest number of common instances. We iteratively apply this process until all topics are mapped to the classes. We evaluate the effectiveness of our method by computing the recognition performance in term of precision, recall, and F1 score. 
 \subsection{Baselines}
We have compared LaHAR to other state-of-the-art methods in HAR. Precisely, \cite{anguita2013public} which was conducted on the same dataset. The authors sampled sequences in a fixed-width sliding window, which is represented by a vector of features such as the mean, standard deviation, correlation, signal magnitude (SMA), etc. A total of 561 features were extracted to describe each subsequence and fed to Support Vector Machine with Gaussian kernels. Using a 10-fold cross-validation procedure, they obtained an overall classification accuracy of about 96\%.\\


 \subsection{Parameter setting}
The extraction of sensory words relies on three main parameters: 1) The sliding window size $p$, which is set to ensure that a sufficient number of sensory words are extracted from the sequences. 2) the overlapping ratio $r$ for subsequences collection which is fixed to $0.5*p$ and 3) the number of characters per channel $v$, which controls the size of the vocabulary. 

To evaluate the impact of the sliding window parameter on the overall result, we conducted our experiments with different values of $p \in \{10, 15, 20, 25, 30, 35\}$. Since the measurements of the dataset are relatively short, we did not consider higher window lengths. Since there is no approved relationship between the size of the vocabulary and the accuracy, we experimented with different values of $v \in \{8, 11, 14, 17, 20, 23, 26, 29\}$ for each $p$ value. \\

\begin{figure*}
\centering
  \begin{subfigure}{0.3\textwidth}
    \includegraphics[width=\textwidth]{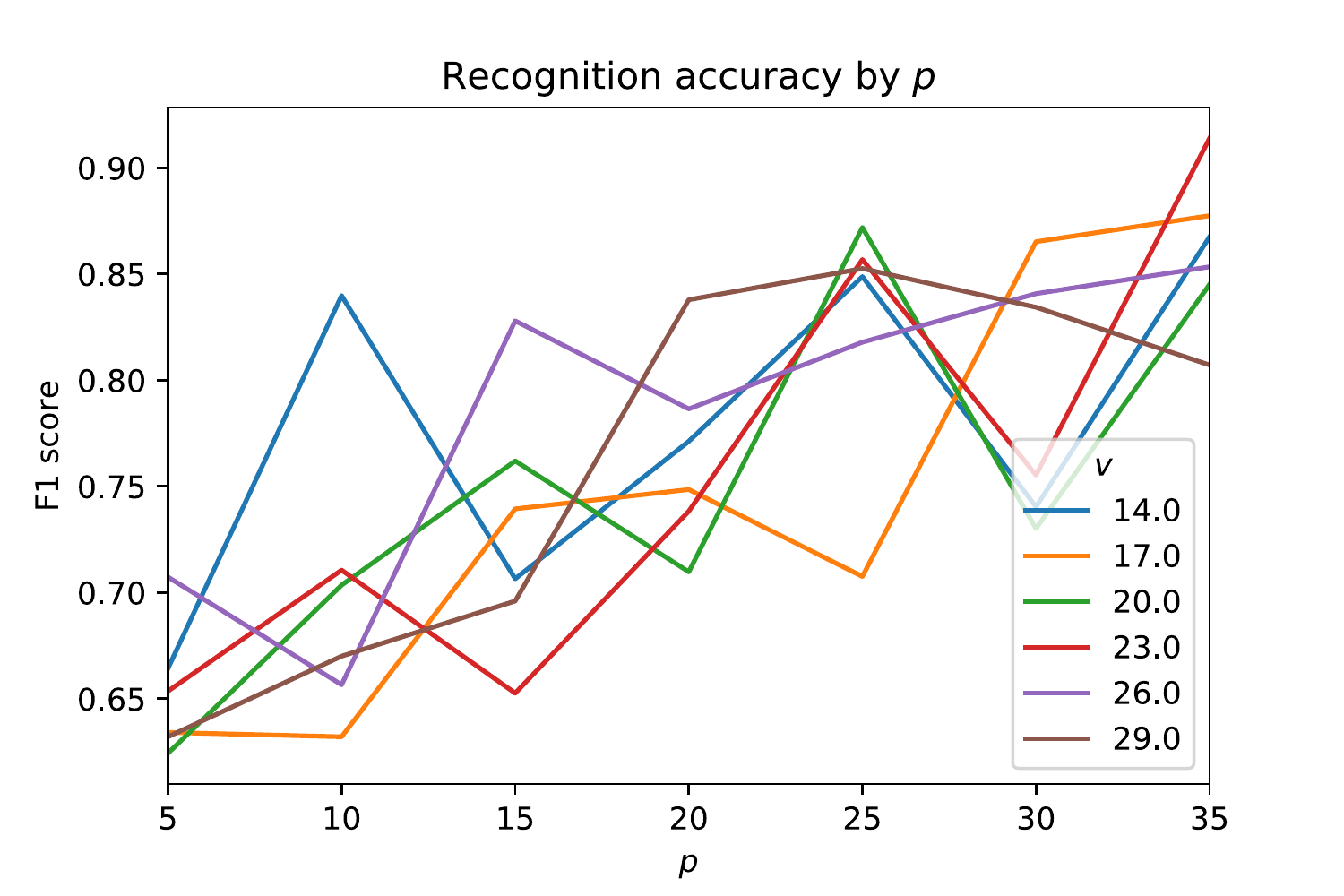}
    \caption{F1 score on training set in terms of $p$}
    \label{fig:awl}
  \end{subfigure}
  \begin{subfigure}{0.3\textwidth}
    \includegraphics[width=\textwidth]{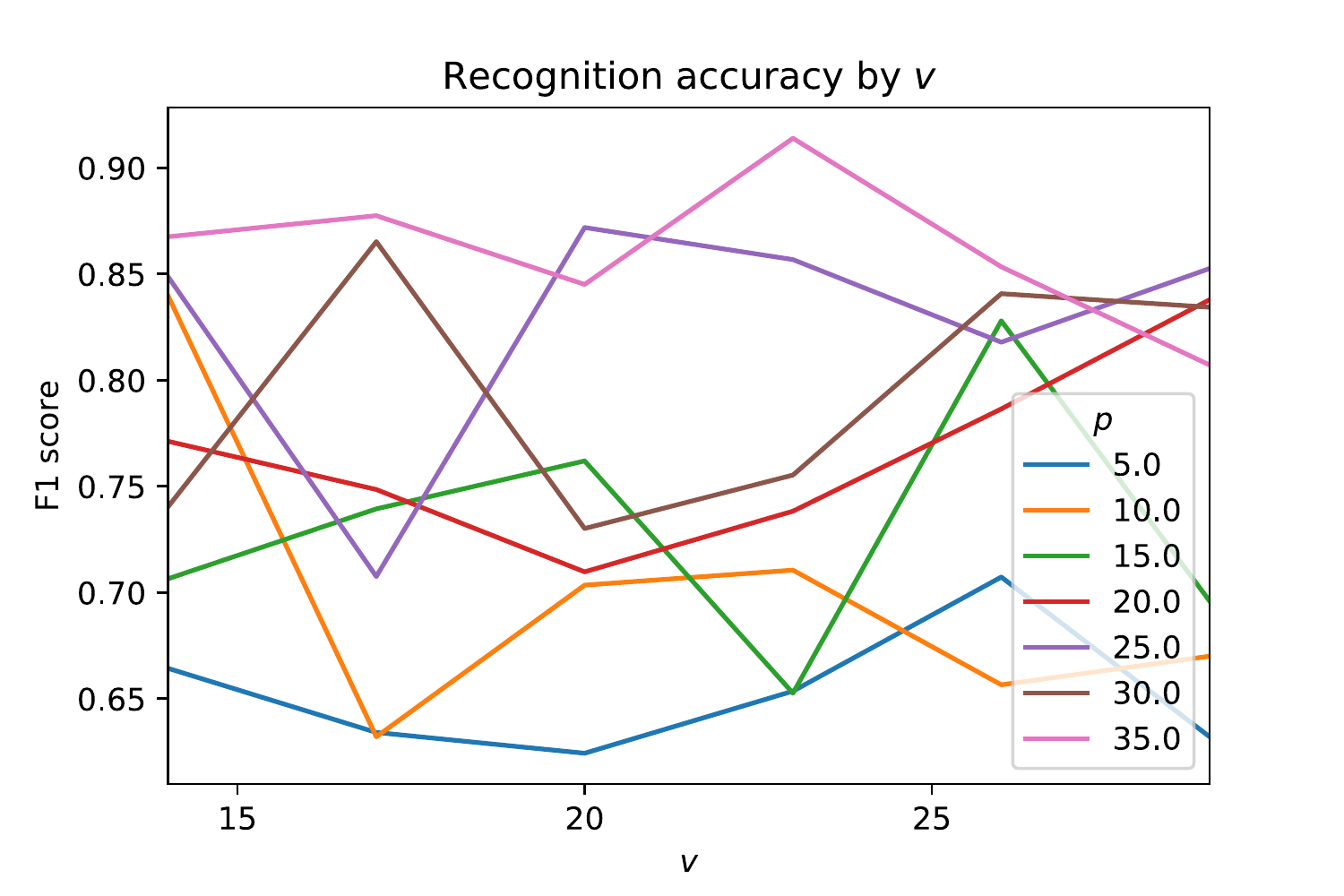}
   \caption{F1 score on training set in terms of $v$}
  \label{fig:anc}
  \end{subfigure}
  \begin{subfigure}{0.3\textwidth}
    \includegraphics[width=\textwidth]{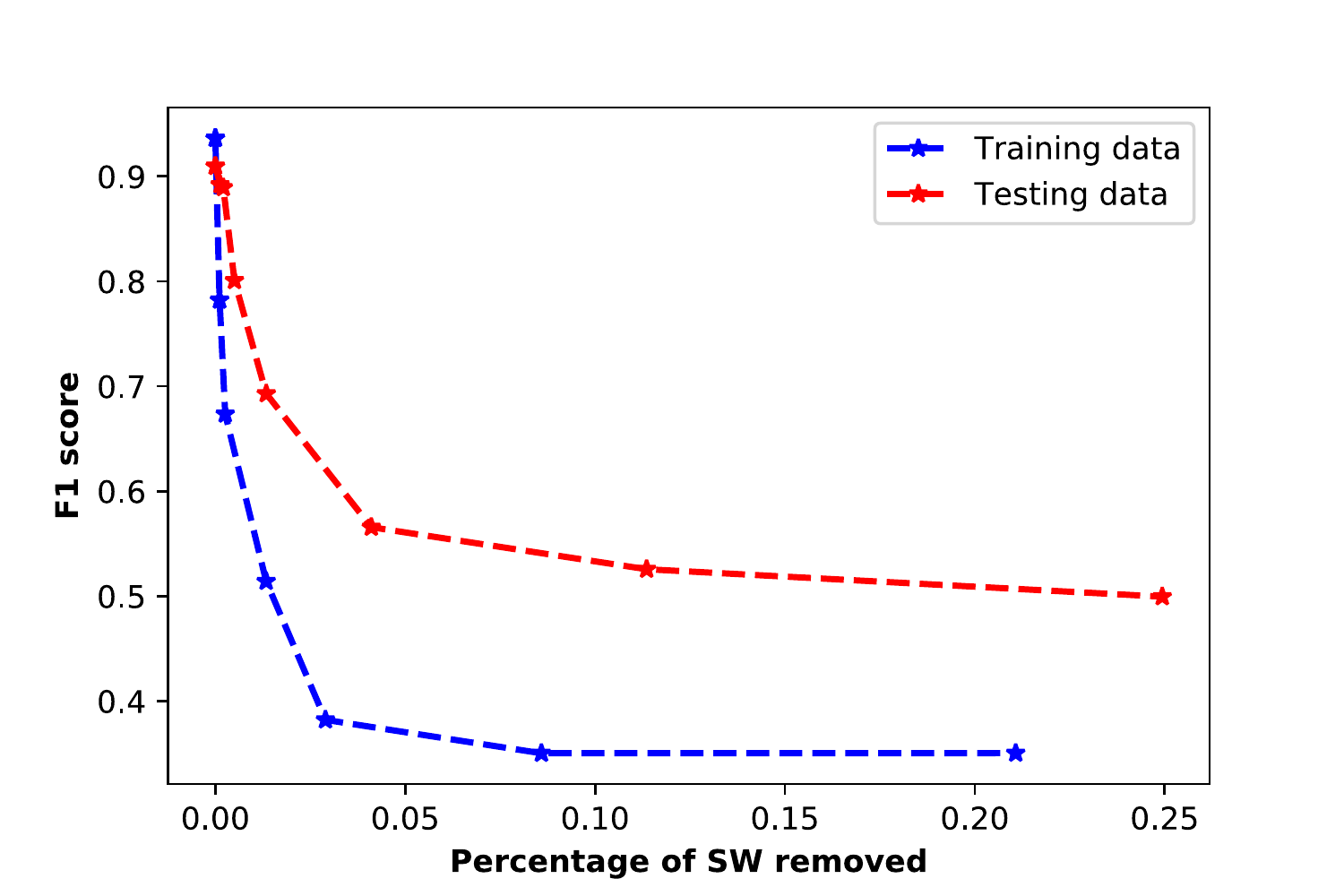}
    \caption{Effect of removing the most frequent SW on the accuracy}
    \label{fig:stop5}
  \end{subfigure}
  \caption{Parameter tuning and prominent sensory words removal}
   \vspace{-0.3cm}
 \end{figure*}

Figures \ref{fig:awl} and \ref{fig:anc} show the distribution of the F1 score obtained from the training set with different window length and number of clusters parameters. For better readability, we did not plot all parameters for the number of clusters. It can be observed that the recognition accuracy increases with high window lengths and number of clusters. Figure \ref{fig:stop5} illustrates the effect of removing the most frequent sensory words on the F1-score. It can be observed that both F1-scores of training and testing dataset drops continuously when the most prominent sensory words are eliminated before running LDA. This obviously suggests all sensory words have to be considered to model activities.

 \begin{figure*}
\centering
  \begin{subfigure}{0.3\textwidth}
    \includegraphics[width=\textwidth]{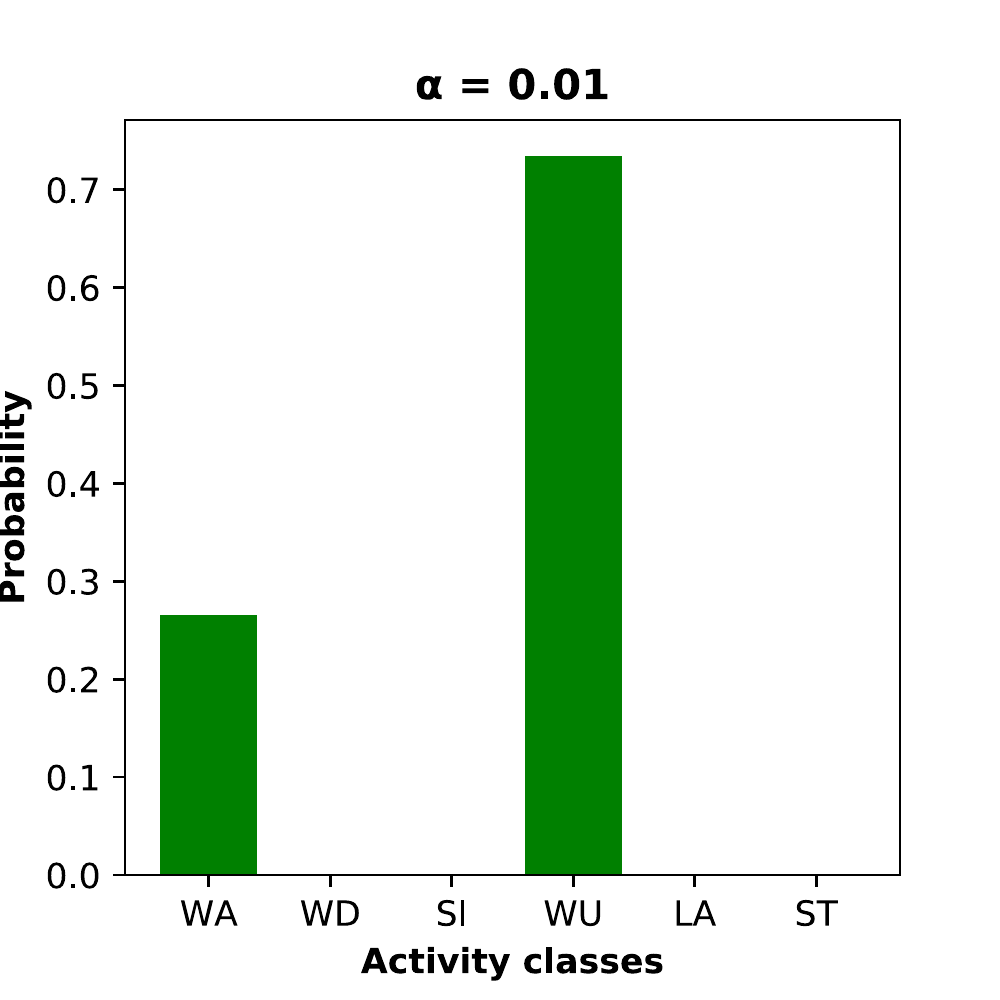}
    \caption{Instance of activity WALKING\_UPSTAIRS}
    \label{fig:wu2c1}
  \end{subfigure}
  \begin{subfigure}{0.3\textwidth}
    \includegraphics[width=\textwidth]{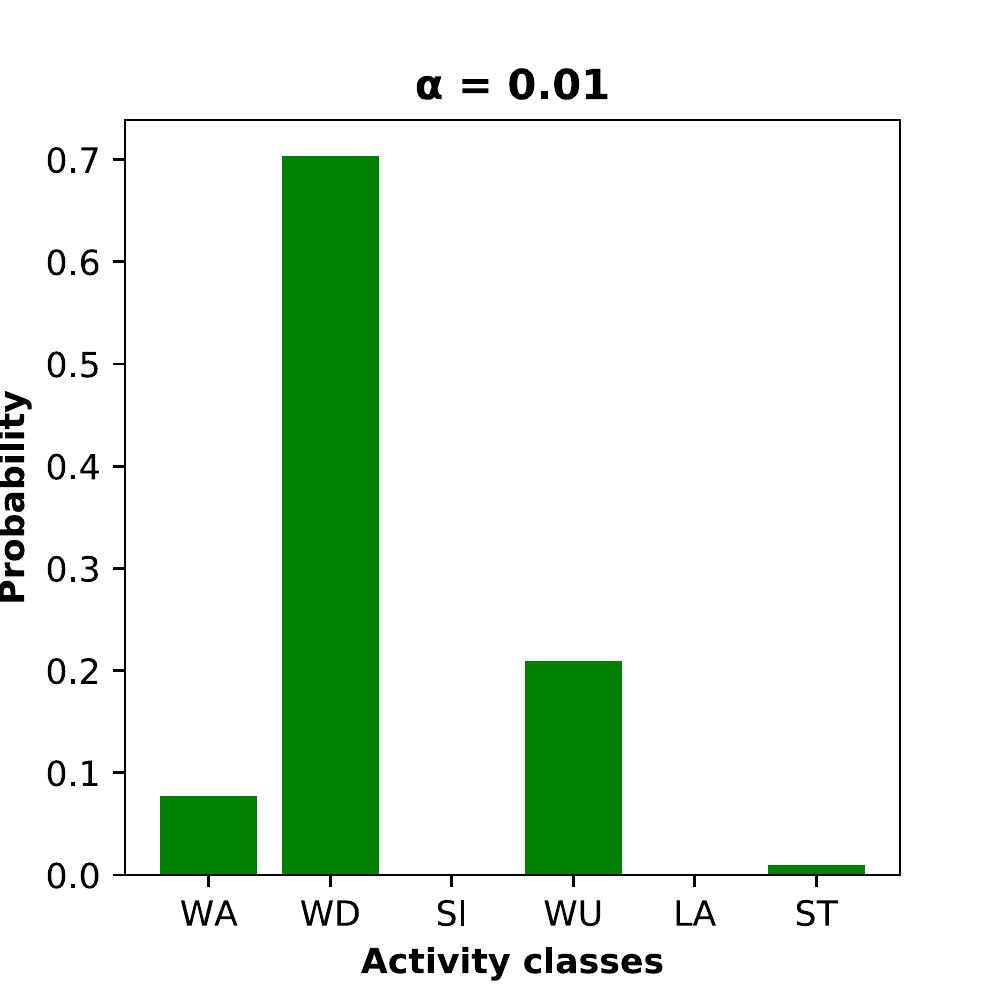}
    \caption{Instance of activity WALKING\_DOWNSTAIRS.}
    \label{fig:si2c1}
  \end{subfigure}
  \begin{subfigure}{0.3\textwidth}
    \includegraphics[width=\textwidth]{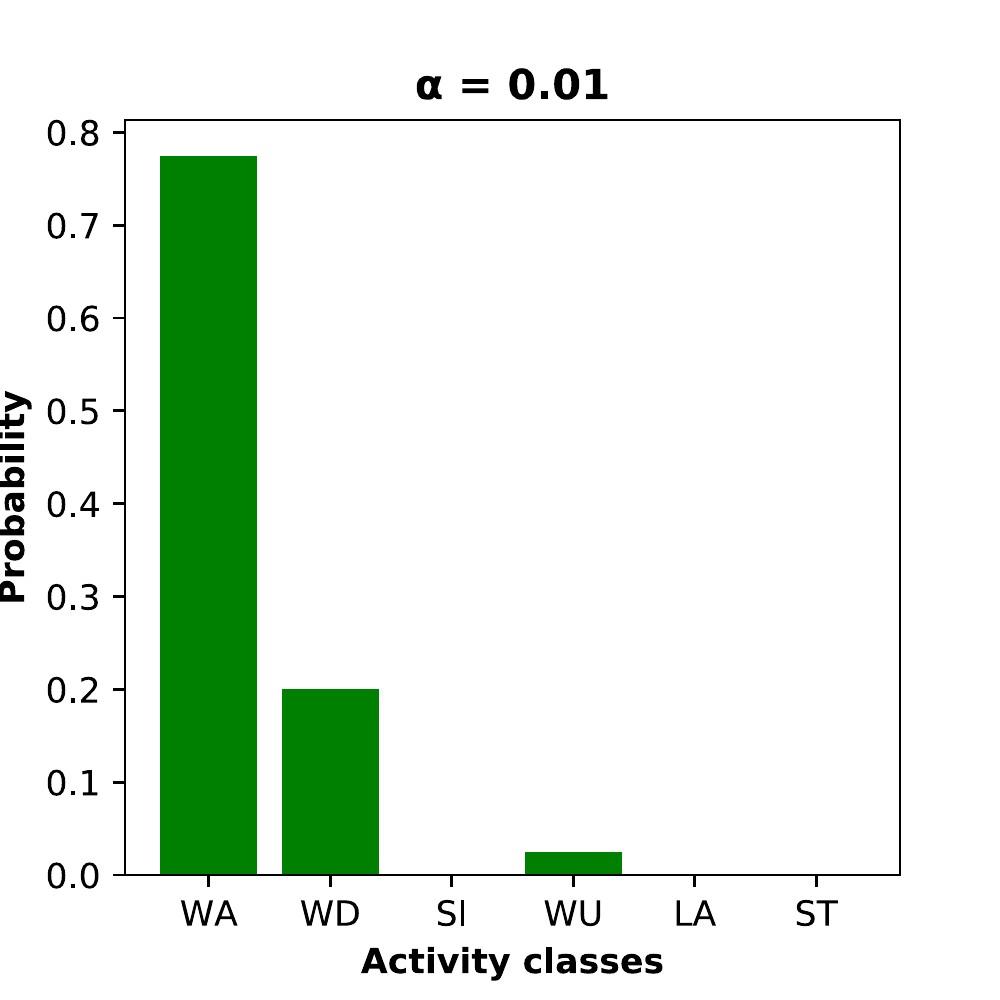}
    \caption{Instance of activity WALKING.}
    \label{fig:wa2c}
  \end{subfigure}
  \caption{Activity class distributions}
  \label{fig:accldist}
    \end{figure*}

\begin{figure}
\centering
  \begin{subfigure}{0.4\textwidth}
    \includegraphics[width=\textwidth]{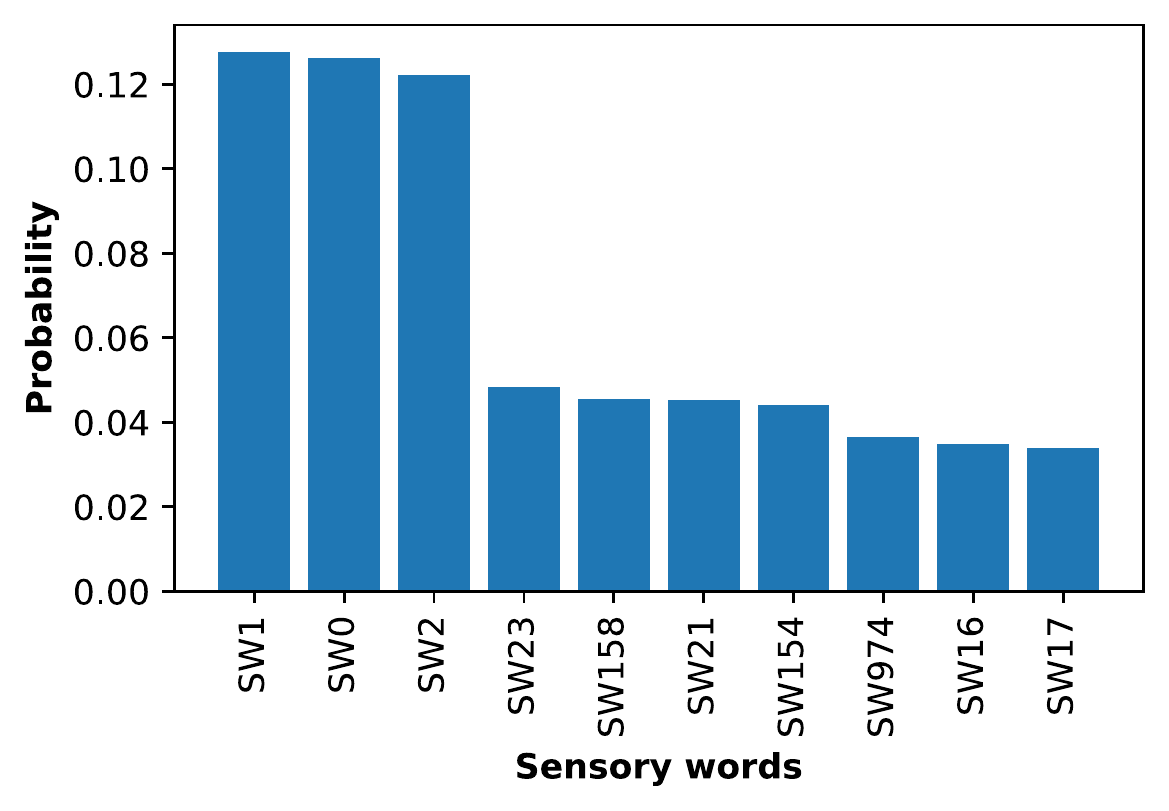}
    \caption{10 most prominent \textit{SWs} in activity SITTING}
   \label{fig:stan}
  \end{subfigure}
  \begin{subfigure}{0.4\textwidth}
    \includegraphics[width=\textwidth]{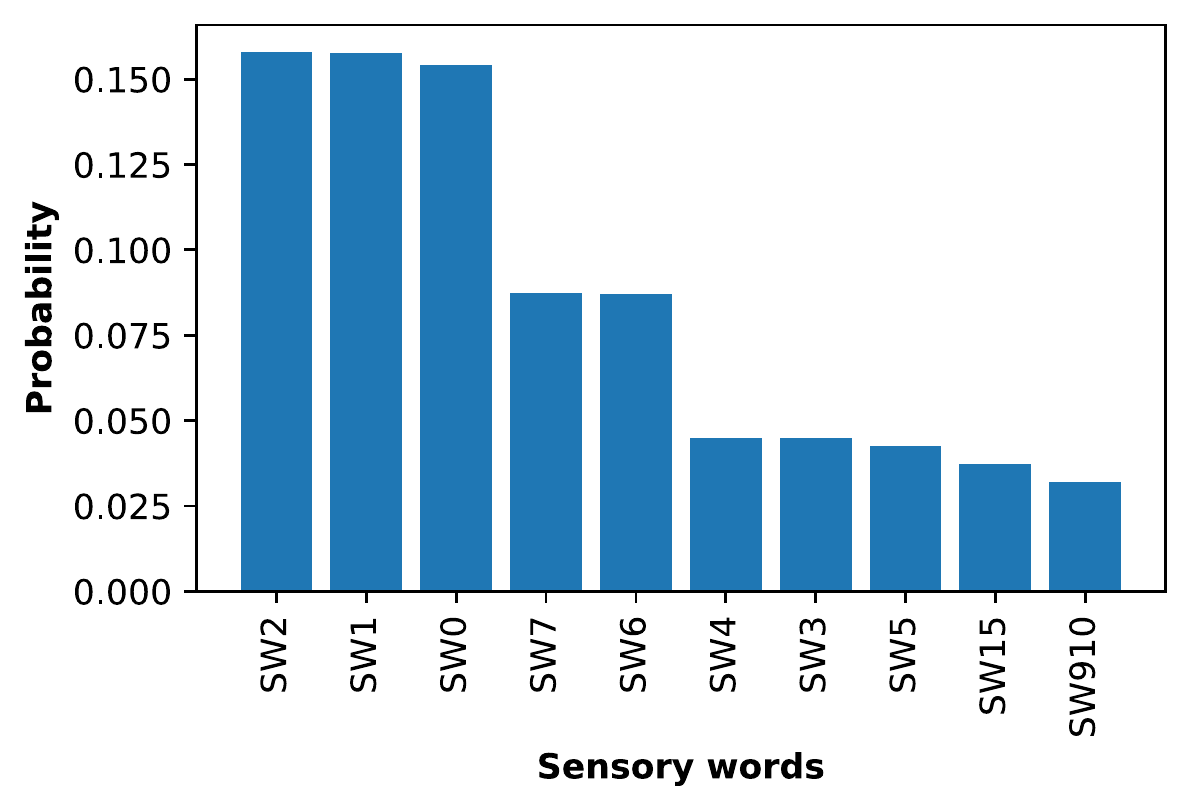}
    \caption{10 most prominent \textit{SWs} in activity STANDING}
    \label{fig:top5m}
  \end{subfigure}
  \caption{Sensory words distributions}
  \vspace{-0.2cm}
\end{figure}



\subsection{Results}
This sub-section reports the results of all experiments conducted in this work. Table~\ref{tab:pr_te2c} provides the recognition accuracy in term of precision, recall, and F1 score for each activity class on the testing dataset. Since the dataset is well-balanced, the macro average was computed for each of the metrics. Despite the fact that LDA is completely unsupervised, it achieved  $91\%$ F1-score which $5\%$ less than the baseline which is a fully supervised approach. Considering the difficulty and cost of annotation, LaHAR achieves a promising result and very close to the result of an expensive supervised approach. The results reveal that LDA can distinguish the activity classes \emph{WA}, \emph{WU}, and \emph{WD} very well from each other, although they are actually very similar. At the same time, the model could not differentiate well between the activities \emph{ST} and \emph{SI}, which in principle are not similar but they are both static activities. This result can be well seen in the confusion matrix presented in Table~\ref{lda_conf}. 

To understand the causes of this misclassification between \emph{ST} and \emph{SI}, Figure~\ref{fig:stan} and Figure~\ref{fig:top5m} show the distribution of sensory words of both activity classes. It can be clearly seen that the two classes share several sensory words. The authors of the dataset considered that the classification error between \emph{SI} and \emph{ST} is due to the position of the sensors while the user was performing the activity \cite{anguita2013public}. Further, \emph{LA} which is also a static activity of 100\%, because the sensor was in a different position. \\

\begin{center}
\begin{table}[ht]
\leftskip=0.5cm
\begin{tabular}{|llll|}
\hline
{Activity classes} &  Precision &  Recall &  F1 score \\
\hline
{WA}            &      0.949 &   0.942 &     0.945 \\
{WU }  &      0.960 &   0.975 &     0.967 \\
{WD} &      0.936 &   0.936 &     0.936 \\
{SI}            &      0.798 &   0.807 &     0.802 \\
{ST}           &      0.827 &   0.806 &     0.816 \\
{LA}             &      0.991 &   1.000 &     0.995 \\
\hline
{\textbf{Mean}}&\textbf{0.910}&\textbf{0.911}&\textbf{0.910}\\
\hline
\end{tabular}
\caption{Results of LDA}
\label{tab:pr_te2c}
\vspace{-0.3cm}
\end{table}
\end{center}

\begin{center}
\begin{table}[ht]
\begin{tabular}{|l|rrrrrr|}
\hline
\backslashbox{Actual}{Predicted} &  WA & WU & WD & SI &  ST & LA \\
\hline
{WA} &473 & 3 &20 & 0 &     0 & 1 \\
{WU} &16 & 459 & 3 & 0 &     0 & 0\\
{WD} &13 &11&393 &0 &     0 &  0\\ 
{SI} &  0 &  1 &  0 &    396 &  89 &     5\\
{ST} &     0 &     3 &  0 &     100 &  429&     0\\
{LA} &     0 &  0 &     0 &     0 &    0 &     537\\\hline
\end{tabular}
\caption{Confusion matrix of LDA}
\label{lda_conf}
 \vspace{-0.8cm}
\end{table}
\end{center}

\subsection{Analysis}
Another goal of applying LDA to HAR data is to validate its capability to captures the mixture components of activity classes. In the UCI HAR dataset, there are no complex activities. However, the activity classes \emph{WA}, \emph{WU}, and \emph{WD} are very similar to each other and hence are overlapped. Figure~\ref{fig:accldist} shows the activity class distributions of three random samples with labels: \emph{WA}, \emph{WU}, and \emph{WD}. The figure demonstrates that even with a low value of $\alpha$, LDA captures the overlapping between the three activity classes. This encourages to apply the model to a dataset with more complex activity classes.

Since LDA has shown great performance on text data especially on large corpora consisting of long documents, we aim to prepare our data to have these properties. For this, Table~\ref{stats} illustrates a statistical comparison between HAR dataset and two text corpora (NIPS and 20 Newsgroups) to which LDA has been successfully applied~\cite{porteous2008}. This comparison is made in terms of 1) the number of instances (documents/data sequences) denoted by $D$, 2) the vocabulary size denoted by $V$, 3) the number of words (i.e. sensory words in our work) per instance, which is denoted by the average in all instances $(B)$ and 3) the number of classes (topics/activity classes) denoted by $K$, as these parameters are known to influence the inference behaviour of LDA \cite{tang2014}. In addition, $N$ represents the total number of words in the collection.  

It can be seen that the values of the parameters, text documents have higher values for \textit{N}, \textit{V}, \textit{B}, and \textit{K} compared to sensory data. However, LDA performed well on sensory data, which implies that the performance of LDA is not determined by the parameters themselves but by the proportion to the size of the vocabulary.

\begin{center}
\begin{table}[ht]
\begin{tabular}{|c|c|c|c|c|c|}
\hline
{Datasets}& $D$ & $N$ & $B$ &$V$ &$ K$ \\ \hline
    NIPS & 1,500 & 1.9 x $10^6$ & 1,267 & 12,419 &400\\ \hline
    20 Newsgroups &11266&1191841&106&18127&20\\ 
	\hline
    UCI HAR& 7352 & 308784 & 63 &5655&6 \\ \hline
	
\end{tabular}
\caption{Statistical comparison between HAR dataset and two text corpora }
\label{stats}
\vspace{-0.6cm}
\end{table}
\end{center}

\section{Conclusion}
\label{conclusion}
In this paper, we presented a novel approach to detect latent patterns in multisensor data acquired for Human Activity Recognition (HAR). To address this problem, we applied Latent Dirichlet Allocation (LDA), which is a topic modelling approach dedicated to cluster large corpora of text documents. To apply LDA on multisensor data, we introduced ``\textit{Sensory Words}'' that are discrete representations of the continuous sequential data. The conducted experiments on a challenging dataset have shown that the results obtained by our unsupervised method are close to those of supervised approaches. This is a useful finding given the growing amount of sensory data available and the difficulty of acquiring annotated training data. In addition, we empirically proved the efficiency of ``\textit{Sensory Words}'' that led to the promising results. We also demonstrated that an activity instance can be represented by a mixture of activities which is beneficial for many HAR tasks.

For future work, in addition to the topics, we will learn their number by using a nonparametric Bayesian method.  In particular, the Hierarchical Dirichlet Process (HDP) that integrates the number of latent topics found in a document collection as part of the posterior distribution of the latent structure, allows unseen documents to generate new topics. This is an attractive property for analyzing growing and changing sensory data. We aim also to apply our method on a more extensive data set with more complex activity classes.

\bibliographystyle{ieeetr}
\bibliography{references.bib}
\clearpage

\end{document}